\documentclass[10pt,twocolumn,letterpaper]{article}

\usepackage[pagenumbers]{cvpr} 

\usepackage{float}
\usepackage{multirow}

\definecolor{cvprblue}{rgb}{0.21,0.49,0.74}
\usepackage[pagebackref,breaklinks,colorlinks,allcolors=cvprblue]{hyperref}

\def\paperID{11644} 
\def\confName{CVPR}
\def\confYear{2025}

\title{Apply Hierarchical-Chain-of-Generation to Complex Attributes Text-to-3D Generation}

\begin{document}

\setlength{\abovedisplayskip}{3pt}
\setlength{\belowdisplayskip}{3pt}
\setlength{\abovedisplayshortskip}{3pt}
\setlength{\belowdisplayshortskip}{3pt}

\twocolumn[{%
\author{Yiming Qin\quad Zhu Xu\quad Yang Liu\thanks{Corresponding author} \\
Wangxuan Institute of Computer Technology, Peking University \\
{\tt\small kevinqym@stu.pku.edu.cn} \quad
{\tt\small xuzhu@stu.pku.edu.cn}  \quad
{\tt\small yangliu@pku.edu.cn} \\
}
\maketitle
\begin{figure}[H]
\hsize=\textwidth
\setlength{\abovecaptionskip}{0.1cm}
\setlength{\belowcaptionskip}{-0.1cm} 
\centering


\vspace{-0.9cm}
\begin{subfigure}{\textwidth}
    \centering
    \includegraphics[width=0.85\textwidth]{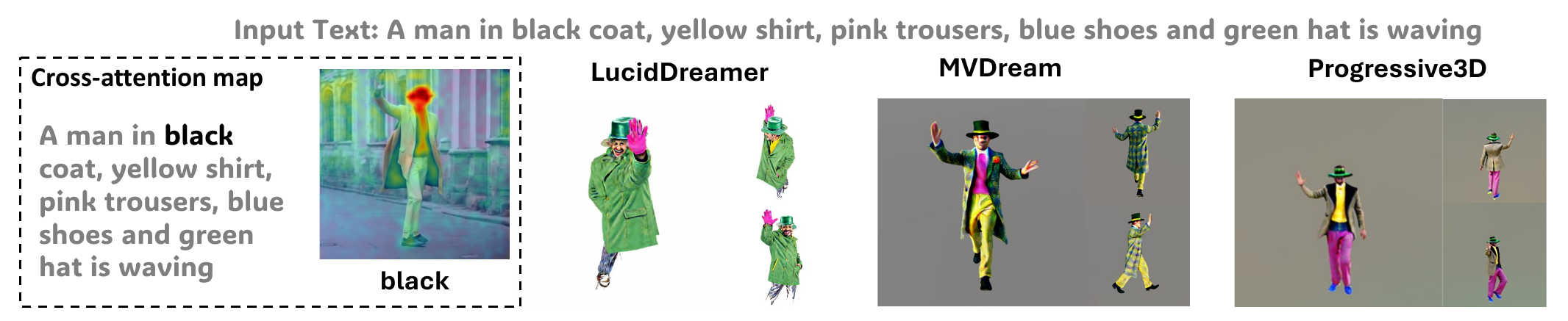} 
    \caption{
    Visualization of cross-attention map. For longer text prompts, 2D Stable Diffusion (SD)~\cite{Rombach2021HighResolutionISSD} fails to accurately associate the word ``black" with the correct spatial location in the generated image. This limitation poses a challenge for methods~\cite{Liang2023LucidDreamerTH, Shi2023MVDreamMD} lifting 2D to 3D using SD effectively.
    }
    \label{fig:teaser1}
\end{subfigure}


\vspace{0.em} 

\begin{subfigure}{\textwidth}
    \centering
    \includegraphics[width=0.8\textwidth]{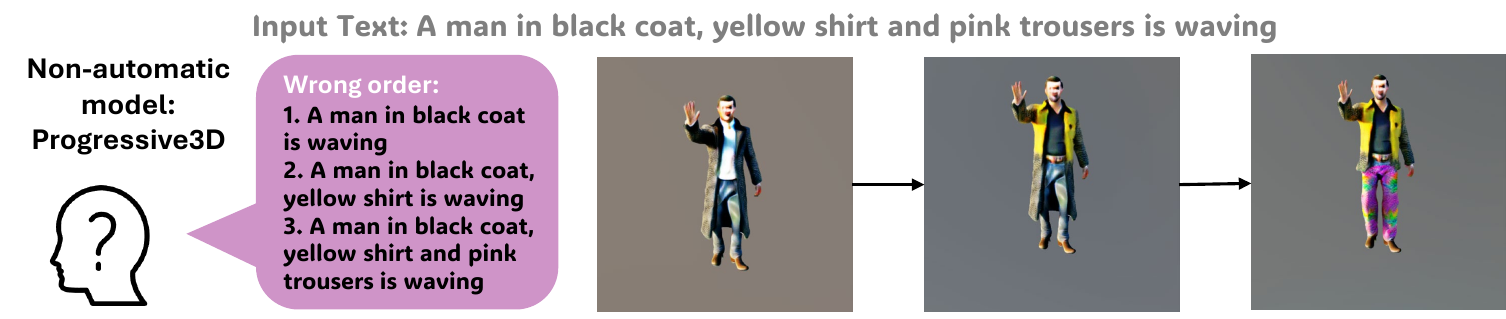} 
    \caption{Some work such as Progressive3D~\cite{Cheng2023Progressive3DPL} that targets 3D generation with complex attributes heavily relies on user-defined bounding boxes and generation order, and imperfect given order results in low-quality results with wrong attributes.}
    \label{fig:teaser2}
\end{subfigure}

\vspace{0.em} 

\begin{subfigure}{\textwidth}
    \centering
    \includegraphics[width=0.85\textwidth]{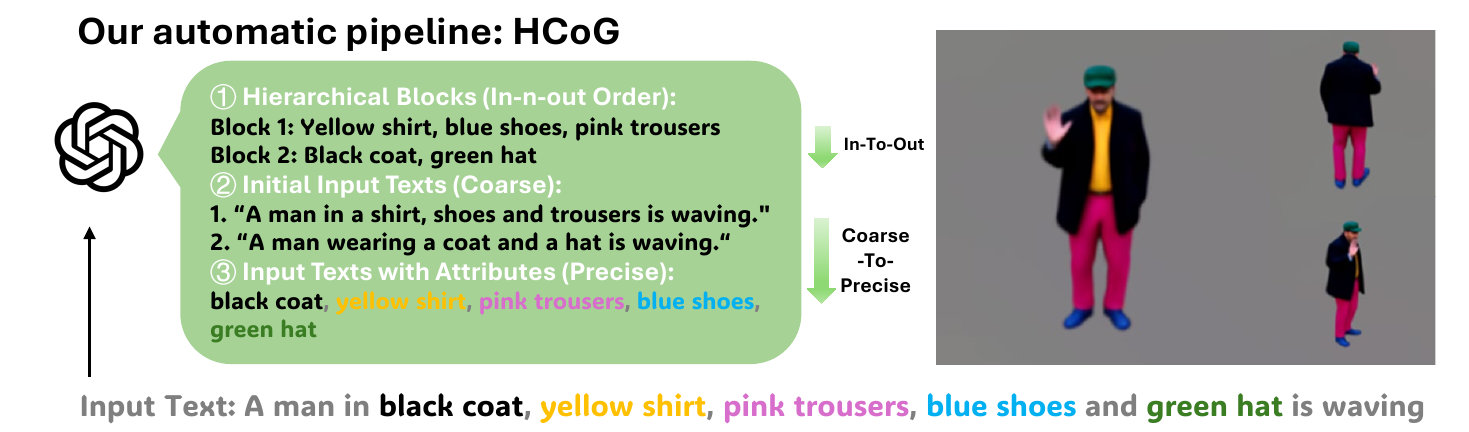} 
    \caption{Our method (HCoG) leverages LLM to generate hierarchical chain of generation, realizing automatic generation of 3D assets with better complex attributes binding capability.}
    \label{fig:teaser3}
\end{subfigure}

\caption{\textbf{The problem of existing work and the example of our method. 
}}
\label{fig:teaser}
\end{figure}
}]

\let\thefootnote\relax\footnotetext{$\ast$ Corresponding author.}

\begin{abstract}

Recent text‑to‑3D models can render high‑quality assets, yet they still stumble on objects with complex attributes. The key obstacles are: (1) existing text-to-3D approaches typically lift text-to-image models to extract semantics via text encoders, while the text encoder exhibits limited comprehension ability for long descriptions, leading to deviated cross-attention focus, subsequently wrong attribute binding in generated results. (2) Occluded object parts demand a disciplined generation order and explicit part disentanglement.
Though some works introduce manual efforts to alleviate the above issues, their quality is unstable and highly reliant on manual information. To tackle above problems, we propose a automated method \textbf{H}ierarchical-\textbf{C}hain-\textbf{o}f-\textbf{G}eneration (\textbf{HCoG}). It leverages a large language model to decompose the long description into blocks representing different object parts, and orders them from inside out according to occlusions, forming a hierarchical chain. Within each block we first coarsely create components, then precisely bind attributes via target‑region localization and corresponding 3D Gaussian kernel optimization. 
Between blocks, we introduce Gaussian Extension and Label Elimination to seamlessly generate new parts by extending new Gaussian kernels, re-assigning semantic labels, and eliminating unnecessary kernels, ensuring that only relevant parts are added without disrupting previously optimized parts.
Experiments confirm that HCoG yields structurally coherent, attribute‑faithful 3D objects with complex attributes. The code is available at \href{https://github.com/Wakals/GASCOL}{https://github.com/Wakals/GASCOL}. 

\end{abstract}
    
\vspace{-1.0em}
\section{Introduction}
\label{sec:intro}

In the field of 3D vision, the development of user-friendly generation of 3D assets with complex attributes has received gaining attention. It allows users to achieve the expected personalized 3D asset generation with few manual efforts. 
For example, Shap-e~\cite{Jun2023ShapEGC} uses the transformer to achieve direct 3D generation. Later works such as DreamFusion~\cite{Poole2022DreamFusionTU} and SJC~\cite{Wang2022ScoreJC} propose to lift prior information in text-to-2D model to 3D, which greatly improves the generalization and detail of the generated results. Since the prior knowledge of 2D diffusion is pervasive, the 2D-based method has stronger generalization and is not limited by the small size of 3D data. Therefore, our work focuses on the method of generating 3D with the help of 2D text-to-image diffusion.

Though significant progress for text-to-3D models, when encountering objects with complex attributes, the results quality of previous approaches like LucidDreamer~\cite{Liang2023LucidDreamerTH} and MVDream~\cite{Shi2023MVDreamMD} still lag behind with attributes deviation, like shown in Fig.~\ref{fig:teaser1}, the cross-attention maps fail to attend to the correct object regions. We claim two primary reasons for such difficulty in complex attributes text-to-3D generation:
Firstly, the widely used CLIP~\cite{Radford2021LearningTVCLIP} text encoder struggles to accurately encode long descriptions, as noted in prior research~\cite{Zhang2024LongCLIPUT}, and may overlook crucial information, leading the cross-attention maps fail to correctly align attribute descriptions with their corresponding image regions. For instance, as shown in Fig.~\ref{fig:teaser1}, the attention of \textbf{black} principally clusters on the head and the shirt, not the coat. Such deviated attentions finally result in wrong attribute binding in generated 3D objects. Secondly, objects with complex attributes naturally exhibit occlusion relationships between different object parts, which requires a reasonable generation order as well as explicit disentangled optimization for these parts to enable structural-coherent and attribute-following results. Some more recent works like Progressive3D~\cite{Cheng2023Progressive3DPL} propose to handle such complex attribute objects by introducing user-defined generation order as well as bounding boxes as guidance. But such manual efforts introduction hinder the generation automation, and the quality of the result heavily relies on user-provided generation order, where an incorrect order of generation will reveal serious consequences as shown in Fig.~\ref{fig:teaser2}. Such failure originates from prioritizing the generation of the external part with less occlusion, during the subsequent generation of the internal part with more occlusion, the surface of the previously generated external part was affected. Thus such manual information-guided generation is also not optimal for complex attributes text-to-3D generation task.

To tackle the above problems and enable high-quality complex attributes text-to-3D generation, we introduce Hierarchical-Chain-of-Generation (HCoG), which generates complext attributes 3D assets in the representation of 3D Gaussian Splatting~\cite{Kerbl20233DGS} (3DGS). Three key designs underpin our HCoG framework: (1) \textbf{Hierarchical Blocks:} we employ a LLM to analyze the text description and decompose the whole object into different parts with shorter descriptions as hierarchical blocks for separate generation. It ensures the quality is not limited by the comprehension ability of text encoder. Further, we propose a \textit{In-n-out order of generation} strategy, which decides the generation order by parts' occlusion relationships, and prioritize to generation inner blocks, which not only can fully expose the inner parts that are occluded for better optimization, but also facilitates the structural integrity of outer parts, yielding more structural coherent results. We also introduce a \textit{generate coarsely-to-precisely} paradigm within each block, which first utilizes coarse-grained attribute-agnostic text for necessary component generation, then applies fine-grained attribute-aware text for detailed attributes editing, improving generation efficiency. (2) \textbf{Part-optimization:} within each block, after necessary components are generated, we target the fine-grained attribute editing for these components. We first adopt \textit{part segmentation} to precisely locate the target region, choosing the corresponding 3D Gaussian kernels for subsequent \textit{fine-grained optimization} to optimize and bind the attributes, during which we introduce MVDream\cite{Shi2023MVDreamMD} and ControlNet\cite{Zhang2023AddingCCControlNet} to facilitate shape control and multi-view consistency for generation. 
(3) \textbf{Gaussian Extension and Label Elimination}: to generate new parts based on previous optimized parts, we first adopt \textit{gaussian extension} to densify the Gaussian kernels to form new parts. To further exclude the affection for previous optimized parts like changing their appearance during such densification process, we introduce \textit{label elimination}, which re-assigns semantic labels for densified new Gaussian kernels, removing unnecessary kernels and only keep kernels that belong to the new part. Notably, our \textbf{Gaussian Extension and Label Elimination} avoid manual input information like user-provided bounding boxes, enabling totally automated generation of HCoG. Besides, HCoG can serve as a plug-and-play generation paradigm for diverse text-to-3D models, showing high scalability. 
Our main contributions are summarized as follows:
\begin{itemize}
    \item We propose Hierarchical-Chain-of-Generation (HCoG), a framework that automates the generation of 3D assets with complex attributes via decomposing the object into hierarchical blocks ordered by occlusion relations for sequential generation.
    \item We propose a coarse-to-fine optimization approach to achieve faithful attribute binding within each hierarchical block. Further, we introduce a Gaussian Extension and Label Elimination strategy between blocks, which eliminates the requirements for manual input guidance for generation and ensures the new-part generation will not affect the optimized parts.
    \item Experiments show that HCoG can automatically generate high-quality 3D assets with complex attributes, especially with strong occlusion relationships, outperforming previous automatic text-to-3D methods. By applying HCoG on different text-to-3D models, we verify its scalability. 
\end{itemize}

\vspace{-1.0em}
\section{Related Work}
\label{sec:relate}

\subsection{Text-to-3D Generation}

The generation of 3D objects has garnered significant attention from researchers, with an increasing number of studies\cite{Wu2016LearningAP, Nichol2022PointEAS, Jun2023ShapEGC, li2023instant3d} focusing on this area. Though great progress, the low quality and scarcity of 3D available data remains a significant challenge to these
3D generation methods. To tackle it, DreamFusion~\cite{Poole2022DreamFusionTU} introduced Score Distillation Sampling (SDS) loss, which distills 2D prior knowledge from a pre-trained diffusion model into the 3D domain, optimizing a 3D representation for each input text. 
Since then, an increasing number of works \cite{Yi2023GaussianDreamerFG,chen2023text, lin2023magic3d, Tsalicoglou2023TextMeshGO, chen2023fantasia3d, Wang2022ScoreJC, Li2023SweetDreamerAG, Liu2023UniDreamUD, Wang2023SteinDreamerVR, Guo2023StableDreamerTN} have been focusing on utilizing some methods to facilitate the process and generate high-fidelity objects. Such as LucidDreamer~\cite{Liang2023LucidDreamerTH}, ProlificDreamer~\cite{wang2023prolificdreamer} and SDS-Bridge~\cite{McAllister2024RethinkingSD} 
modify SDS loss function to generate higher-quality objects. 
Among these methods, Progressive3D~\cite{Cheng2023Progressive3DPL} have made a contribution on the generation of 3D assets with complex attributes. However, it needs plenty of manual work and will fail if users make mistakes about the order of generation and the location of bounding boxes when there is obvious occlusion in target generated asset.

\subsection{3D Editing}

When we need more customized 3D assets and 3D data or want to optimize some 3D data, 3D Editing is an essential tool. However, there are certain challenges in editing 3D data precisely. EditNeRF~\cite{Liu2021EditingCR} is an early proposal for editing 3D data, which uses coarse 2D user scribbles to edit the neural radiance field. After this work, a large plenty of effort has been put into editing the neural radiance field. SINE~\cite{Bao2023SINESI}, TextDeformer~\cite{Gao2023TextDeformerGM}, CLIP-NeRF~\cite{Wang2021CLIPNeRFTD}, ED-NeRF~\cite{Park2023EDNeRFET} propose one-stage method to edit neural radiance field based on the given text or the reference image. On editing 3D Gaussian Splatting, the pioneering work GaussianEditor~\cite{Chen2023GaussianEditorSA}, Gaussctrl~\cite{Wu2024GaussCtrlMC} and ~\cite{Luo20243DGE} edit 3D Gaussian Splatting using text prompt or referred images. All these works directly edit 3D scenes based on simple input texts or images. Different from these works, our work aims to decompose complex long input text and leverage the insight of editing to optimize test-to-3D assets with careful consideration of optimization order
, which ensures the accurate generation of objects with complex attributes.

\begin{figure*}[!t]
    \centering
    \setlength{\abovecaptionskip}{0.1cm}
    \setlength{\belowcaptionskip}{-0.6cm} 
    \includegraphics[width=0.92\linewidth]{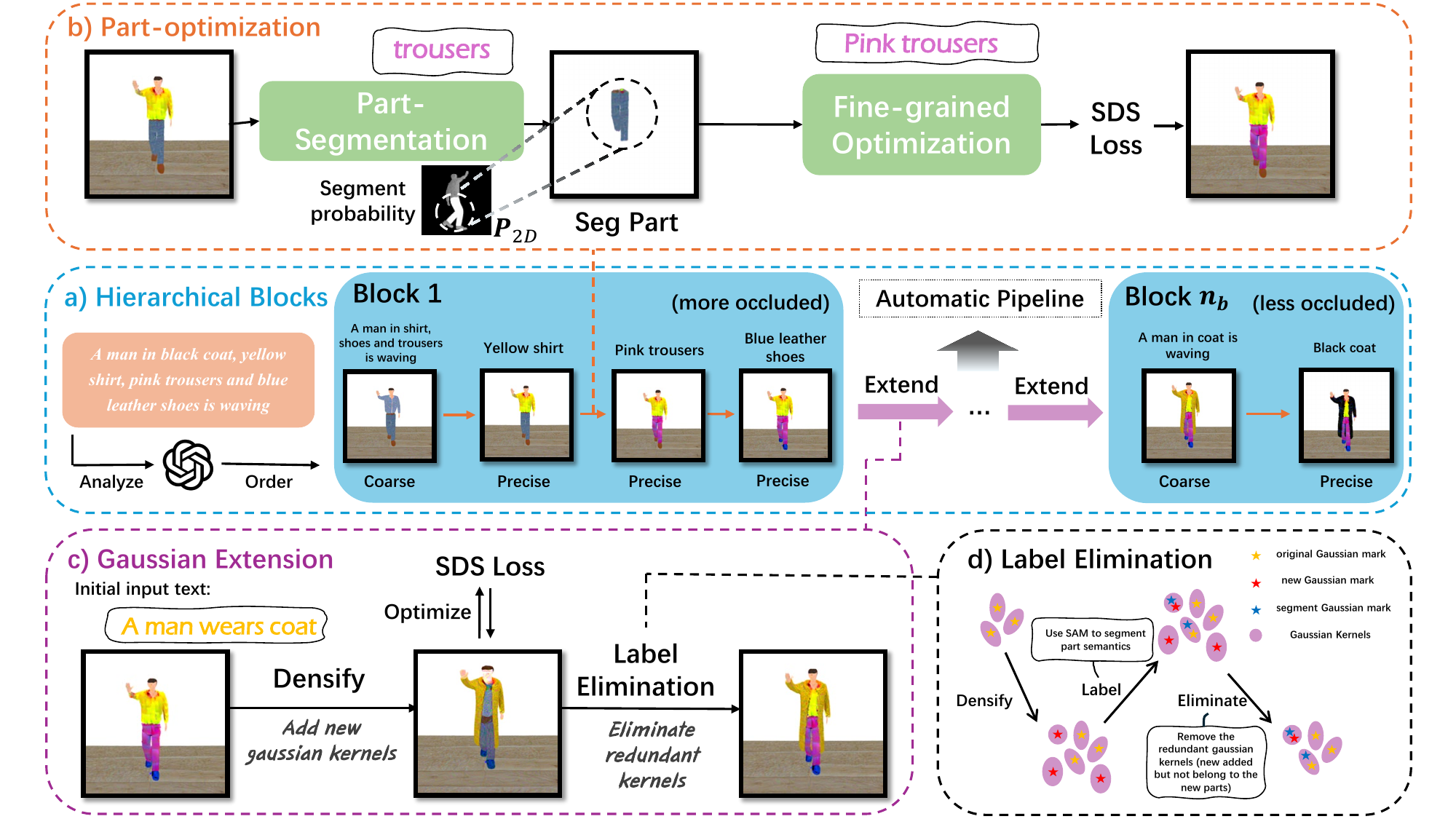}
    \caption{\textbf{Overview of Hierarchical-Chain-of-Generation.} a) In the Hierarchical Blocks stage, LLM analyzes the input text and based on the order from more occlusion to less occlusion, creating the order of generation. b) Part-optimization is applied to the parts in blocks, using Lang-SAM~\cite{luca2024langSAM} to segment specific parts and utilizing MVDream~\cite{Shi2023MVDreamMD} and ControlNet~\cite{Zhang2023AddingCCControlNet} in fine-grained optimization stage to enable corresponding attributes binding for each part with shape and multi-view consistency. c) Gaussian Extension is applied between blocks, extending new parts for the next block. d) Label Elimination aims to generate new
    parts by extending new Gaussian kernels (red-star-marked), re-assigning semantic labels (blue-star-marked), and eliminating unnecessary kernels finally, ensuring that only relevant parts are generated without disrupting previously optimized parts.
    }
    \label{fig:method}
\end{figure*}

\vspace{-0.3em}
\section{Preliminaty Knowledge}
\label{sec:preliminary}

\subsection{3D Gaussian Splatting}
\label{subsec:3dgs}

Gaussian Splatting utilizes a set of 3D Gaussian kernels to fit the 3D scene or object, serving as one powerful 3D representation with high-quality. Formally, Gaussian kernels can be parameterized as $\theta$. For each $\theta_i = \{\mathbf{x}_i, \mathbf{s}_i, \mathbf{q}_i, \mathbf{c}_i, \alpha_i\}$, where $\mathbf{x}_i \in \mathbb{R}^3$ represents the coordinate of the center of the i-th Gaussian in Cartesian coordinate system, $\mathbf{s}_i \in \mathbb{R}^3$ represents the scaling size, $\mathbf{q}_i \in \mathbb{R}^4$ is the rotation of the i-th Gaussian which is represented as a quaternion, $\mathbf{c}_i \in \mathbb{R}^3$ contains the RGB of this Gaussian kernel and the $\alpha_i \in \mathbb{R}$ is the opacity value. The whole space is served as tile list to be projected onto the screen plane by a sample camera, and the color $\mathbf{C}(\mathbf{p})$ of each point $\mathbf{p}$ on the projection screen is calculated with the formula:$\mathbf{C}(\mathbf{p}) = \sum_{i\in \mathcal{N}} \mathbf{c}_i \alpha_i^\prime \prod_{j=1}^{i-1} (1 - \alpha_j^\prime),$
where $\alpha_i^\prime = \alpha_i e^{-\frac{1}{2}(\mathbf{p} - \mathbf{x}_i)^T \Sigma_i^{-1} (\mathbf{p} - \mathbf{x}_i)}$, $\Sigma_i$ is the covariance of the i-th Gaussian which can be arrived by $\mathbf{s}_i$ and $\mathbf{q}_i$, and $\mathcal{N}$ denotes the number of Gaussians in this tile. Since 3D Gaussian Splatting is a display expression, we can use the characteristics of the display expression to map each kernel to the corresponding part of the 3D asset, making it easier to optimize a certain part separately.

\subsection{Score Distillation Sampling (SDS)}
\label{subsec:sds}

Score Distillation Sampling proposed in DreamFusion~\cite{Poole2022DreamFusionTU} aims to distill the prior knowledge in 2D diffusion models for 3D generation. A key advantage of this technology is its independence from 3D data. In the absence of 3D data, SDS technology demonstrates enhanced generalization capabilities and can produce a wider range of diverse results, making it especially beneficial for users seeking to generate personalized 3D assets with complex attributes. In this paradigm, the parameter of the 3D scene $\theta$ is denoted as differentiable image parameterization. Then, after rendering image $\mathbf{I}_{RGB}$ according to a given camera pose $\mathbf{p}_{c}$, a Gaussian noise $\epsilon(0, I)$ is added onto the image and passed into the diffusion model $\phi$. With the predicted noise $\epsilon_{\phi}$ at timestep $t$, SDS loss optimize the 3D scene parameter $\theta$ by calculating the difference between added noise $\epsilon(0, I)$ and the predicted noise $\epsilon_{\phi}$, which is formulated as,
\begin{equation}
    \nabla_{\theta} \mathcal{L}_{\text{SDS}} = \mathbb{E}_{\epsilon,t} \left[
    w(t) \left( \epsilon_{\phi} (\mathbf{I}_{RGB}; y(\mathbf{p}_{c}), t) - \epsilon \right) \frac{\partial \mathbf{I}_{RGB}}{\partial \theta}
    \right]
\label{eq:SDS}
\end{equation}
where $y(\mathbf{p}_{camera})$ is the text prompt related to the camera pose $\mathbf{p}_{c}$ and $w(t)$ is a weight function. To speed up the process of backpropagation, SDS loss simply skips the grad of U-net.
\vspace{-0.4em}
\section{Method}
\label{sec:method}
\vspace{-0.4em}

The framework of Hierarchical-Chain-of-Generation is shown in Fig.~\ref{fig:method}, which comprises three main designs. We first adopt \textbf{Hierarchical Blocks} (detailed in Sec.~\ref{sec:method-1}) to analyze the text description and decompose the whole generation into sequential blocks. To yield more structural coherent results, in-n-out order strategy are further introduced to generate these blocks from inner ones to outer ones according to the occlusion relationships between them. Then within each block, we propose \textbf{Part Optimization} (detailed in Sec.~\ref{sec:method-2}) to first coarsely generate necessary components, then precisely edit and bind corresponding attributes through target region localization and fine-grained 3D gaussian kernel optimization, yielding faithfully and accurate attribute binding for each object part. Between blocks, \textbf{Gaussian Extension and Label Elimination} (detailed in Sec.~\ref{sec:method-3}) is introduced to generate new parts, which firstly extend new Gaussian kernels through kernel densification, then re-assign semantic labels for newly generated kernels, and eliminate redundant ones according to labels, which only eliminate the manual efforts to make whole generation automatic, but also avoid negative affection like appearance changing in previous optimized parts.

\subsection{Hierarchical Blocks}
\label{sec:method-1}

As shown in Fig.~\ref{fig:method}, 
when faced with complex 3D objects with multiple constituent parts and distinct attributes, it is highly probable that certain parts may occlude others, resulting in only partial visibility of some elements from any given viewpoint, thereby confusing 2D diffusion model and complicating the generation of 3D assets. To address the issue of attribute binding and ensure a fully automated pipeline, we propose Hierarchical Blocks, which consist of two key design principles: (1)
\textbf{In-n-out order of generation:} We first extract all object parts from the long complicated input text and organize them into hierarchical layers according to their occlusion relationships. The most occluded parts are placed in the initial layer and generated first, while the least occluded parts are assigned to the final layer and generated last. Parts that do not occlude one another are grouped within the same layer, allowing for parallel generation. (2) \textbf{Generate coarsely-to-precisely:} Within each hierarchical block, all constituent parts are first generated by the initial input text, and then are refined as attribute binding by the fine-grained input texts. Specifically, the initial input text includes only the parts in this block, 
such as \textbf{``a man in shirt, shoes and trousers is waving"} shown in session a) of Fig.~\ref{fig:method}, is coarsely initial input text, omitting detailed attributes and generating parts in this block. Progressively, \textbf{``yellow shirt"} is the precise optimization input text, enhancing the attribute ``yellow" bound to ``shirt".

\textbf{In-n-out order of generation. }Adopting our generation order ensures more structurally coherent results. Specifically, heavily occluded parts are often entirely or partially invisible in the final rendered image, making post-generation optimization nearly impossible. Therefore, it is crucial that these occluded parts are generated and optimized before the outer parts are introduced. By ensuring that the outer layers remain ungenerated during this process, the system can fully expose the occluded components, allowing them to be accurately reconstructed and optimized. Moreover, the structural integrity of the outer parts depends largely on the shape of the inner parts. Thus, a hierarchical generation strategy, which prioritizes inner (more occluded) parts before outer (less occluded) ones, ensures better global consistency in the final 3D asset. For parts that do not occlude each other, their generation order remains flexible and they can be grouped within the same hierarchical block to enable parallel processing.

\textbf{Generate coarsely-to-precisely. }The generation of all parts in the same block is based on generating all new parts coarsely and editing precisely. In each block, the initial input text is responsible for generating all parts associated with that block simultaneously. Since 3D generation is computationally intensive, producing multiple components in a single forward pass significantly improves efficiency compared to sequential generation. To optimize the generative model’s performance, the initial input text excludes attribute-level details, as minimizing textual complexity enhances the model’s ability to learn high-level structural information. Once the coarse representation of all parts is established, subsequent fine-grained refinements are performed, where each part’s attributes are explicitly defined and serially incorporated into the generation process. 
An illustrative example is presented in Fig.~\ref{fig:teaser3}.
The hierarchical structure specifies the grouping of parts into different blocks. The initial input text for each block captures a high-level structural outline of all the parts within that block, such as ``A man in a shirt, shoes and trousers is waving" contains all ``shirt", ``shoes" and ``trousers" in this block. Meanwhile, it also provides precise descriptions of each part along with its attributes, such as ``yellow shirt", to bind attribute to the corresponding part.

To make this process totally automated, we use a large language model to help us create \textbf{Hierarchical Blocks}. Since many objects or parts in daily life have occlusion relationships, the large language model (LLM) has prior knowledge of occlusion relationships, assisting us to analyze the occlusion relationships of various parts in long complex input texts, and further providing the order of generation based on occlusion relationships. In addition, the LLM is capable of extracting various parts and corresponding attributes and providing simplified initial input texts, which is shown in Fig.~\ref{fig:teaser3}.

\begin{figure*}[t]
    \centering
    \setlength{\abovecaptionskip}{0.2cm}
    \setlength{\belowcaptionskip}{-0.6cm} 
    \includegraphics[width=0.9\linewidth]{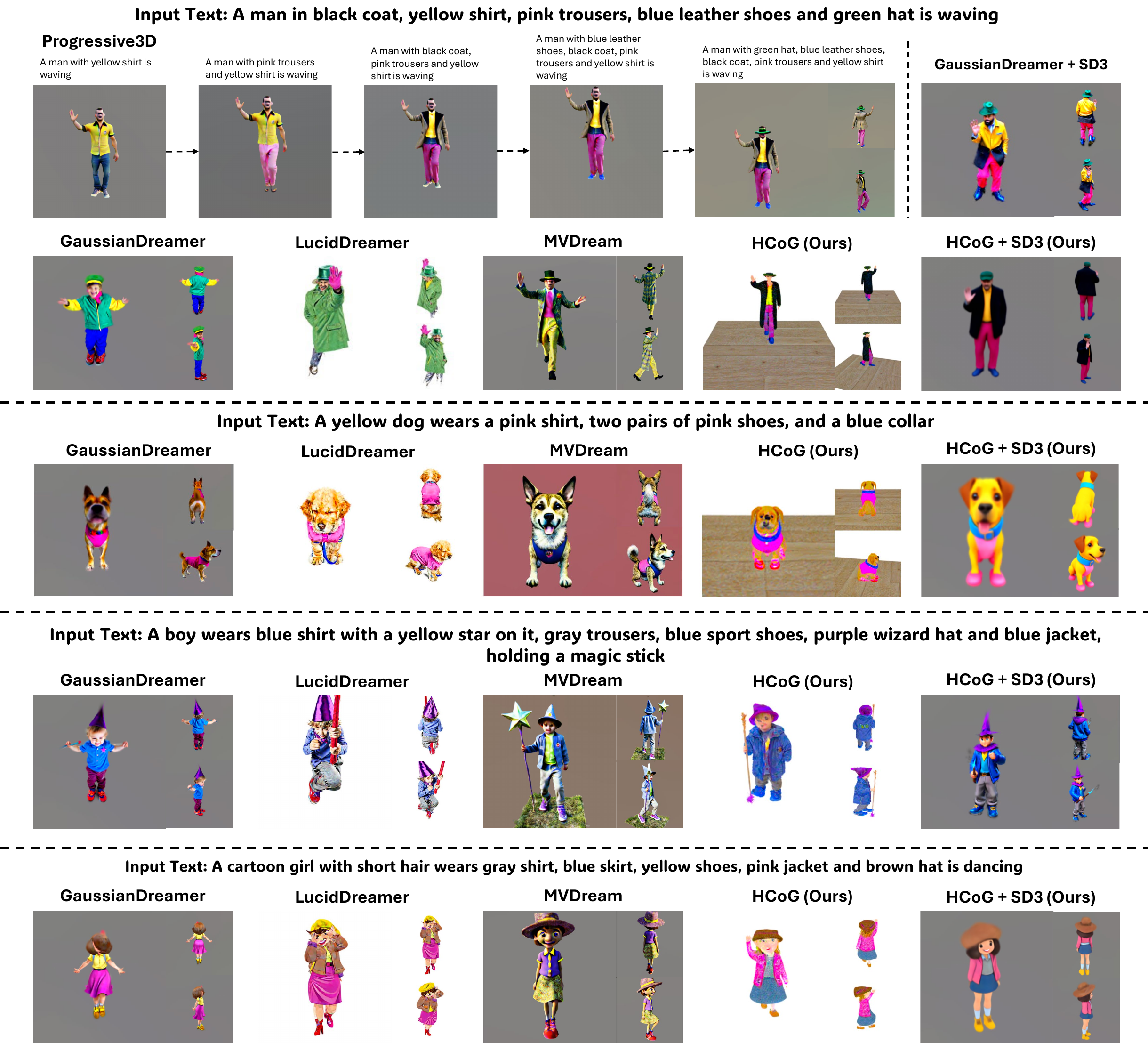}
    \caption{\textbf{Visual comparison with other methods.} We compare our method with other well performed text-to-3D methods~\cite{Yi2023GaussianDreamerFG, Liang2023LucidDreamerTH, Shi2023MVDreamMD}, Progressive3D~\cite{Cheng2023Progressive3DPL} which heavily relies on user-defined generation order and bounding boxes, and Stable Diffusion v3~\cite{Rombach2021HighResolutionISSD} which is a more powerful backend.}
    \label{fig:result}
\end{figure*}

\subsection{Part-optimization}
\label{sec:method-2}

Within each hierarchical block, multiple parts coexist without occluding one another. Once the coarse structures are generated, each part undergoes precise optimization, which first requires accurate localization and segmentation. To achieve this, we employ SAM~\cite{Kirillov2023SegmentASAM} to segment 2D rendered images and lift the 2D to 3D. Subsequently, the fine-grained optimization is based on the framework of MVDream~\cite{Shi2023MVDreamMD} which ensures multi-view consistency, adopting SDS loss to optimize 3D assets which are formulated as Eq.~\ref{eq:SDS}.

\textbf{Part Segmentation} is the first stage, aiming to segment certain part that needs optimization and using lang-SAM~\cite{luca2024langSAM} to supervise the segmentation of the target part. Specifically, we bind a new property on each Gaussian kernel, labeled $p_{seg}$, randomly initialized, which acts as a binary label, indicating the kernel's possibility of belonging to the target part. The goal of this stage is to optimize $p_{seg}$ of all kernels so that it converges to a point where the kernels belonging to the certain part have higher $p_{seg}$. For each time the 3D asset is rendered in a sampled camera pose, the binary label $p_{seg}$ is rendered as a 2D tensor $\mathbf{P}_{2D}$, which means the current binary label renders as 2D tensor with segmentation possibility and needed optimization. In the meantime, an RGB image $\mathbf{I}_{RGB}$ is rendered based on the same camera pose. Then, we apply lang-SAM\cite{luca2024langSAM} to the rendered image to obtain a segmentation ground-truth $\mathbf{P}_{\mathbf{I}_{RGB}}$. The segment loss is formulated as:
\begin{equation}
    \mathcal{L}_{segment} = \text{CrossEntropy}(\mathbf{P}_{2D}, \mathbf{P}_{\mathbf{I}_{RGB}})
\end{equation}
Through iterations of training, we yield the converaged binary classification labels $p_{seg}$ to group the Gaussian kernels belonging to the target part. As shown in session b) of Fig.~\ref{fig:method}, after multi-iteration, the Segment probability converges on the trousers and this part is segmented out.

\textbf{Fine-grained Optimization} is the second step to optimize the target part, aiming to bind correct attributes to new corresponding parts. For each part, we only allow the Gaussian kernels belonging to the target part to receive gradients while fixing the other Gaussian kernels. During the optimization process, since the quality of generated assets may not be optimal at the early stage and some parts show differences in sizes and shapes from prior knowledge of diffusion models, we further introduce to combine ControlNet~\cite{Zhang2023AddingCCControlNet} to provide the shape prior to the generated assets to diffusion, which corrects the error caused by the gap between the generated Gaussian Splatting and the real Gaussian Splatting data. Therefore, the segmented part is fed in to ControlNet~\cite{Zhang2023AddingCCControlNet} for better shape consistency, applied SDS loss:
\begin{equation}
    \mathcal{L}_{optim} = \mathcal{L}_{\text{SDS}}^{\text{ControlNet}} + \mathcal{L}_{\text{SDS}}^{\text{MVDream}}
\end{equation}
\vspace{-0.2em}
where $\mathcal{L}_{\text{SDS}}^{\text{ControlNet}}$ is the $\mathcal{L}_{\text{SDS}}$ from ControlNet~\cite{Zhang2023AddingCCControlNet} and $\mathcal{L}_{\text{SDS}}^{\text{MVDream}}$ is from MVDream~\cite{Shi2023MVDreamMD}. The formula of $\mathcal{L}_{\text{SDS}}$ is referred as Eq.~\ref{eq:SDS}.

\subsection{Gaussian Extension and Label Elimination}
\label{sec:method-3}
\vspace{-0.1cm}
In order to reduce manual efforts like precise user-defined bounding boxes and not change the appearance of previous optimized parts, we propose the Gaussian Extension operation to generate new parts without changing the original Gaussian kernels and the Label Elimination to ensure new parts not change the appearance of previous parts which have been optimized.

\textbf{Gaussian Extension.} The goal of the Gaussian Extension operation is to generate the parts of the next block while preserving the previous parts, and it does not depend on the bounding boxes defined by the user. As shown in Fig~\ref{fig:method} Part c), we densify the original 3DGS and get some new Gaussian kernels. For each new Gaussian kernel, it is generated from a Gaussian kernel in the original 3DGS. The properties of the new Gaussian kernel are copied from the original Gaussian Kernel, except for the position. The position of new kernel $\mathbf{x}_{new}$ will be sampled on the distribution of the original Gaussian kernel $N(\mathbf{x}_{origin}, \mathbf{s}_{origin})$ and a small random perturbation will be added, which can be formulated as:
\begin{equation}
    \mathbf{x}_{new} = \mathbf{x}_{sample} + \mathbf{x}_{perturb}
\end{equation}
where $\mathbf{x}_{origin}$ and $\mathbf{s}_{origin}$ denote the center and the scale of the original gaussian kernel, $\mathbf{x}_{sample} \sim N(\mathbf{x}_{origin}, \mathbf{s}_{origin})$ denotes sampling a point on the distribution of original Gaussian kernel and $\mathbf{x}_{perturb} \sim N(0, \mathbf{\epsilon})$ means random perturbation and $\epsilon$ is a tiny noise covariance.
We fix all the original 3DGS and only allow the new 3DGS to accept gradients, thus to ensure preserve the original parts. Then, we use the simplified next-block text provided by LLM to optimize all 3DGS via SDS loss, so that all the parts of the next block can be optimised.

\textbf{Label Elimination.} Although the Gaussian Extension method does not require the user to define the precise position of the bounding boxes, it leads to attribute deviation within previous parts, which is shown in Sec.~\ref{sec:ablation}. 
The reason is, during Gaussian Extension stage, the extended kernels may be attached the the surface of the previous optimized parts. Therefore, to preserve the appearance of these optimized parts, we propose Label Elimination, a concise and effective way to remove the influence of the parts in the previous block and only keep the new parts in the next block.

Specifically, as shown in Fig.~\ref{fig:method}, purple circles representing kernels and colorful stars representing marks, before the Gaussian Extension step, all the original gaussian kernels are orange-star-marked.
After Gaussian Extension method, the densified Gaussian kernels are red-star-marked. After the optimization, we apply SAM~\cite{Kirillov2023SegmentASAM} to segment the parts that are novel in this new block. All Gaussian kernels that are segmented as new parts are blue-star-marked. After this operation, the blue-star-marked kernels contain the parts we want to add, while the red-star-marked kernels 
are actually not belong to the parts we want to add, which means they are redundant. Subsequently, we only need to eliminate the red-star-marked kernels to get a harmonious 3D asset with new parts without negatively affecting the original optimized parts.

Besides, because different Gaussian kernels are
required as the size varies for each part, Label Elimination can help remove redundant Gaussian kernels so that during the process of Gaussian Extension, we only need to add a fixed number of Gaussians.

\begin{table}[t]
\centering
\setlength{\abovecaptionskip}{0.1cm}
\setlength{\belowcaptionskip}{-0.5cm} 
\setlength{\tabcolsep}{2pt}
\resizebox{1.0\linewidth}{!}{\begin{tabular}{lccccc}
\hline
            & GSD~\cite{Yi2023GaussianDreamerFG} & MVD~\cite{Shi2023MVDreamMD} & Pro3D*~\cite{Cheng2023Progressive3DPL} & Ours & Ours+SD3 \\ \hline
BLIP-VQA    & 0.4919 & 0.5519 & 0.6553 & \underline{0.7295} & \textbf{0.8055} \\
CLIP-Score  & 30.709 & 31.132 & 30.451 & \underline{31.998} & \textbf{33.189} \\ \hline
\end{tabular}}
\caption{\textbf{Quantitative Comparison.} Where GSD means GaussianDreamer, MVD denotes MVDream, and Pro3D* represents Progressive3D. Our method outperforms other methods.}
\label{table:quantitative_comparison}
\end{table}

\vspace{-0.2em}
\section{Experiments}
\label{sec:exp}
\subsection{Implementation Details}
Our hierarchical chain-of-generation (HCoG) framework is designed to be compatible with various backbones. In this paper, we implement HCoG on both GALA3D~\cite{Zhou2024GALA3DTT} based on Stable Diffusion v2.1 and GaussianDreamer~\cite{Yi2023GaussianDreamerFG} equipped with the advanced text-to-image model Stable Diffusion v3.
We use GPT-4o as the large language model. During the process of \textbf{Part Segmbentation}, a threshold is needed to identify which part a Gaussian kernel belongs to and is set to 0.9. $p_{seg}$ is trained for 200 iterations where the learning rate is set to 0.05. The camera's sampling radius is set to the range of the scene in a spherical coordinate system, while vertical angles are sampled uniformly from $-45^\circ$ to $45^\circ$ and horizontal angles are sampled uniformly from $360^\circ$. 
In the Process of \textbf{Extend}, the random perturbation $\epsilon$ is set to 0.01. 


\vspace{-0.3em}
\subsection{Qualitative Comparisons}
The comparison of visualized results with other methods is shown in Fig.~\ref{fig:result}. We choose some well-performed text-to-3D methods~\cite{Yi2023GaussianDreamerFG, Liang2023LucidDreamerTH, Shi2023MVDreamMD, Cheng2023Progressive3DPL} to compare with our method (Ours). Besides, we upgrade the 2D diffusion model used as SDS guidance to more advanced Stable Diffusion v3, and compare GaussianDreamer~\cite{Yi2023GaussianDreamerFG} with our method (Ours+SD3), both equipped with Stable Diffusion v3.  
For testing, we selected challenging examples featuring complex parts and attributes, including those with distinct occlusion relationships, to evaluate the model's performance in complex attribute text-to-3D generation.



\textbf{Compare with Progressive3D~\cite{Cheng2023Progressive3DPL}.} As shown in the first row of Fig.~\ref{fig:result}, Progressive3D~\cite{Cheng2023Progressive3DPL} performs generally satisfactorily and shows competitive results when the input text is complex. However, it needs plenty of manual effort. Users is requested to define the generation order and bounding boxes carefully, which makes it less practical. Besides, if the order is not proper, the results may be a bad crash. As shown in Fig.~\ref{fig:teaser2}, generating black coat first and yellow shirt later will dye the coat yellow. In comparison, our effort has the ability to automatically decide the optimization order and generate 3D assets without any manual effort.

\textbf{Compare with SOTA text-to-3D methods.} As shown in Fig~\ref{fig:result}, we perform visual comparison of our method with GaussianDreamer~\cite{Yi2023GaussianDreamerFG}, MVDream~\cite{Shi2023MVDreamMD} and LuciDreamer~\cite{Liang2023LucidDreamerTH}, which are all equipped on the same version of Stable Diffusion as ours. In these cases, admittedly, these methods are able to generate delicate and high-quality results, while they are unable to tackle the task of 3D generation with such complex attributes. They usually bind the parts with wrong attributes in the input text, such as binding the star on the magic stick instead of the shirt, while sometimes missing some parts, such as the two pairs of pink shoes in that dog. Meanwhile, our method have the ability to tackle these situations.

\textbf{Compare with GaussianDreamer~\cite{Yi2023GaussianDreamerFG} with Stable Diffusion v3.} When upgrading the 2D diffusion model to the more advanced, recent Stable Diffusion v3 (as shown in the right-most column), we observe that GaussianDreamer~\cite{Yi2023GaussianDreamerFG} demonstrates some improvement but remains inadequate in attribute binding. It bind ``yellow" to ``coat" by mistake. In contrast, when equipping Stable Diffusion v3 to HCoG, our method enables more fine-grained details and higher-quality outputs while maintaining the correctness of attribute binding. Note that the order of operations remains critical when upgrading 2D diffusion model to Stable Diffusion v3. Additional ablation studies can be found in Sec.~\ref{sec:ablation}.



\begin{table}[t]
\centering
\setlength{\abovecaptionskip}{0.1cm}
\setlength{\belowcaptionskip}{-0.4cm} 
\resizebox{0.8\linewidth}{!}{\begin{tabular}{lccc}
\hline
            & Inverse order & Random order & Ours \\ \hline
HCoG  & 0.5023 & 0.5961 & \textbf{0.7295} \\
+SD3      & 0.6363 & 0.7033 & \textbf{0.8055} \\ \hline
\end{tabular}}
\caption{\textbf{Ablation of Generation order on BLIP-VQA.}}
\vspace{-0.05cm}
\label{table:ablation_order}
\end{table}

\begin{figure}[t]
    \centering
    \setlength{\abovecaptionskip}{0.1cm}
    \setlength{\belowcaptionskip}{-0.2cm} 
    \includegraphics[width=0.9\linewidth]{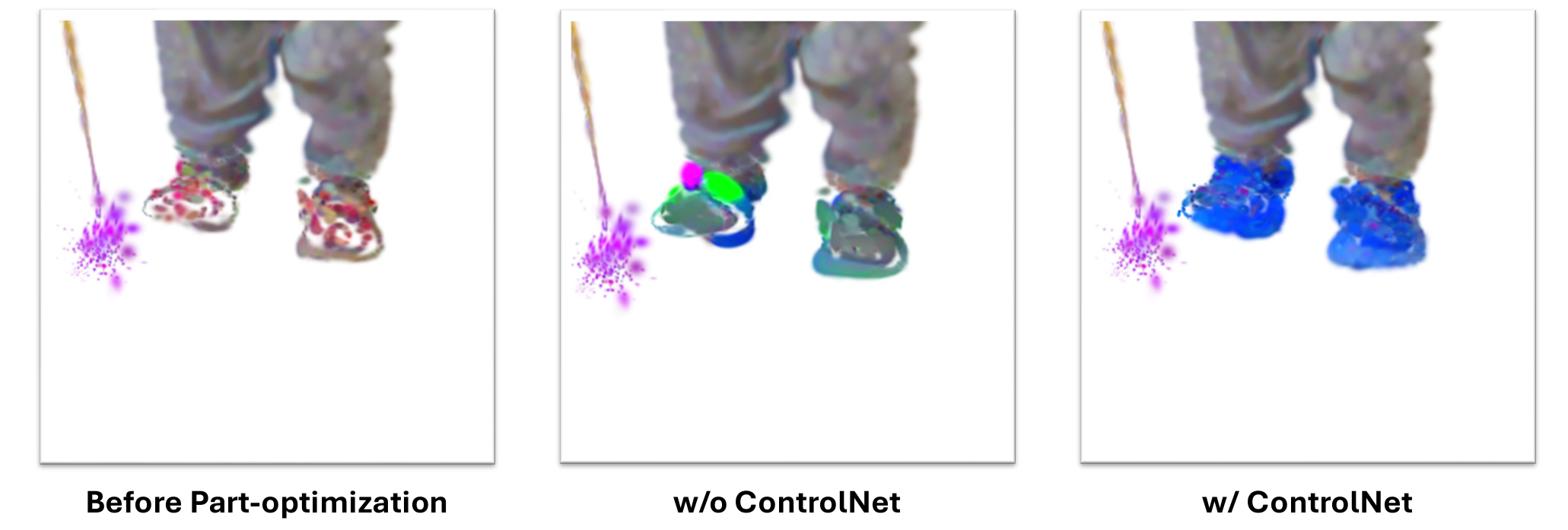}
    \caption{\textbf{Ablation study of ControlNet. Input text: blue sports shoes.} Without shape control, the diffusion model will give wrong guidance and the result will be bad.}
    \label{fig:ablation-control}
\end{figure}

\begin{figure}[t]
    \centering
    \setlength{\abovecaptionskip}{0.1cm}
    \setlength{\belowcaptionskip}{-0.5cm} 
    \includegraphics[width=0.9\linewidth]{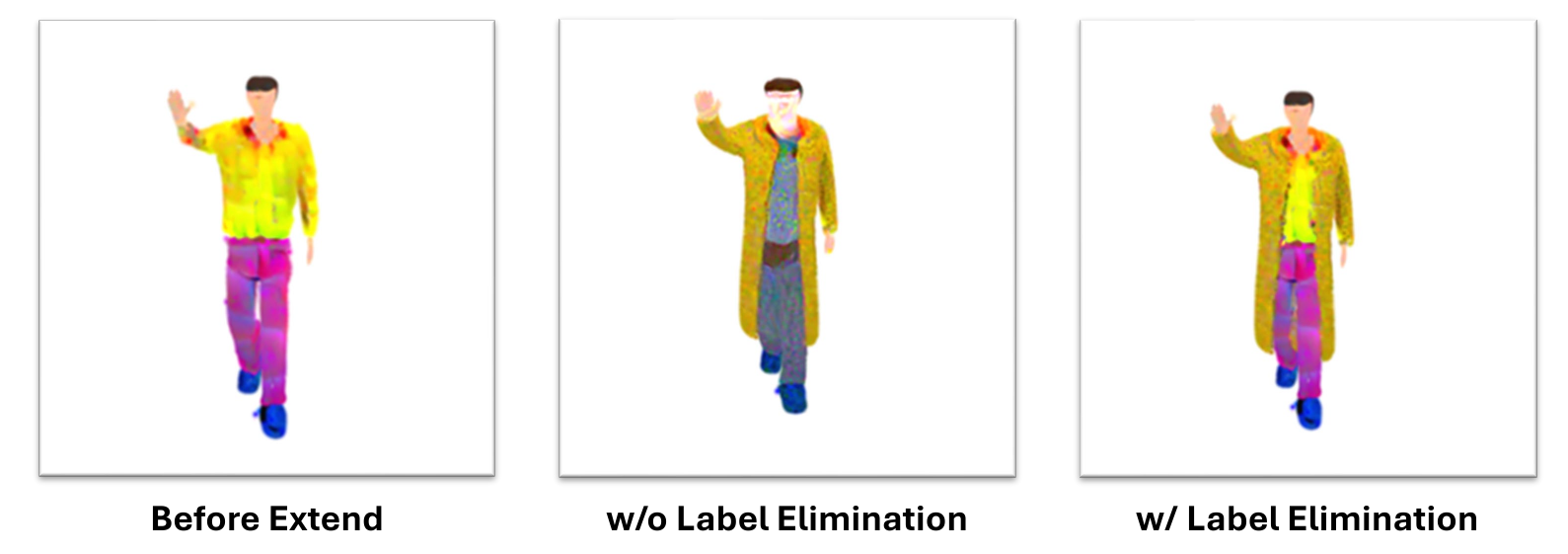}
    \caption{\textbf{Ablation study of Label Elimination. Previous optimized input text: A man in yellow shirt, pink trousers and blue leather shoes is waving. Next input text: A man in coat is waving.} Without Label Elimination, when generating new part \textbf{coat}, the optimized parts like \textbf{yellow shirt}, \textbf{pink trousers} and \textbf{blue leather shoes} are changed.}
    \label{fig:ablation-LE}
\end{figure}

\vspace{-0.4em}
\subsection{Quantitative Comparison}
\vspace{-0.4em}
\label{sec:exp-quan}

In order to quantitatively compare the results, we adopted the BLIP-VQA~\cite{Li2022BLIPBL} method proposed in T2I-CompBench~\cite{Huang2023T2ICompBenchAC} and CLIP~\cite{Radford2021LearningTVCLIP} similarity score to evaluate the quality of the generated results. However, according to the T2I-CompBench~\cite{Huang2023T2ICompBenchAC}, the BLIP-VQA scores of each part are multiplied to get the total score. However, 
due to our set-up of ``3D assets with complex attributes" being challenging, we empirically observed that the original BLIP-VQA score of most methods is zero. Therefore, in order to enable a more intuitive comparison, we changed the total score to the average of the scores of each part, and finally obtained the results shown in Tab.~\ref{table:quantitative_comparison}. As can be seen, our method is better than many previous text-to-3D methods ~\cite{Yi2023GaussianDreamerFG, Shi2023MVDreamMD}, and also outperforms Progressive3D~\cite{Cheng2023Progressive3DPL} which targets for complex attributes binding by 7.42\% on BLIP-VQA and by 1.547 on CLIP-score, achieving the best score. The other advantage is that our method needs no manual effort.
Besides, when we take Stable Diffusion v3 as our backend, the score achieves 0.8055 on BLIP-vQA and 33.189 on CLIP-score, which implies stronger 2D diffusion model provides more reliable guidance.

\vspace{-0.4em}
\subsection{Ablation Experiments}
\label{sec:ablation}
\vspace{-0.4em}
We conduct an ablation study to evaluate the effectiveness of the order of generation, ControlNet~\cite{Zhang2023AddingCCControlNet} and Label Elimination respectively.

\textbf{Order of generation.} 
We conduct an ablation experiment on the order of generation and the results is shown in Tab.~\ref{table:ablation_order}, which reveals that generating in order from the severely obscured to the lightest obscured ensures each part can be well optimized. Even equipped with more advanced SD3, the order of generation is still crucial in generating high quality results. 


\textbf{ControlNet.} We conduct experiments with and without ControlNet~\cite{Zhang2023AddingCCControlNet} to verify its effectiveness. As shown in Fig.~\ref{fig:ablation-control}, ControlNet provides shape and size information to the diffusion model, ensuring stable optimization. Empirically, ControlNet~\cite{Zhang2023AddingCCControlNet} is essential for Part-optimization, as it bridges the gap between real and generated data for Gaussian Splatting, preventing issues with distorted shapes and sizes that confuse the diffusion model.



\textbf{Label Elimination.} 
We conduct experiments with and without Label Elimination to verify its effectiveness. As shown in
Fig.~\ref{fig:ablation-LE}, which indicates that Label Elimination is the key design to generate new parts without interfering with the previous optimized parts by removing redundant kernels attached to the surface of previous optimized parts. 


\vspace{-0.4em}
\section{Conclusion}
\label{conclusion}
\vspace{-0.4em}

We present a method Hierarchical-Chain-of-Generation (HCoG) that targets complex attributes text-to-3D generation task. It utilizes a LLM to analyze the input text description, decomposes the object into hierarchical blocks with different object parts to generate them sequentially with the order decided by their occlusion relationships. Within each block, a coarse-to-fine optimization process is conducted to faithfully bind attributes for each part. Between blocks, gaussian kernels extension and label elimination are proposed to smoothly generate new parts without disrupting previously optimized ones.The entire pipeline is fully automated, minimizing manual effort and enhancing user-friendliness. Experiments demonstrate the effectiveness and scalability of our method.

\section*{Acknowledgment}

This work was supported by the grants from the Beijing Natural Science Foundation 4252040 and  National Natural Science Foundation of China 62372014.

{
    \small
    \bibliographystyle{ieeenat_fullname}
    \bibliography{main}
}

\clearpage
\appendix
\setcounter{page}{1}

\maketitlesupplementary
\let\thefootnote\relax\footnotetext{$\ast$ Corresponding author.}

In this supplementary material, we provide additional information about prompt engineering and conduct additional experiments on the CSP-100 test dataset provided in ~\cite{Cheng2023Progressive3DPL}. We begin with providing prompt engineering details in Sec.~\ref{sec:prompt}. Subsequently, we conduct: (1) Additional experiments on CSP-100 test dataset (Sec.~\ref{sec:suppl-csp}).]; (2) Analysis of Large Language Model (Sec.~\ref{sec:analysisofLLM}); (3) Experiments using Long-CLIP~\cite{Zhang2024LongCLIPUT} (Sec.~\ref{sec:supply-longclip}).

\section{Prompt Engineering}
\label{sec:prompt}

The structure of HCoG relies on the capabilities of the large language model (LLMs). Therefore, we provide the details of our prompt engineering strategy. To avoid unnecessary complexity in the main text, we present only the overall approach. First, we instruct the LLM on the task it is expected to perform specifically, decomposing a relatively complex input text into a sequence of shorter texts in an ordered manner.
We then explicitly describe the ``in-out" generation order to the LLM, clarifying that the target decomposition should follow a progression from the most heavily occluded (inner) instances to the least occluded (outer) ones. This step ensures that the LLM understands the intended generation sequence before performing the task. Next, we offer one or two illustrative examples to help the LLM better understand the task. Finally, we supply the target input text, prompting the LLM to perform the actual decomposition. The detailed and referred prompts can be found in the examples provided in our code repository.

\begin{figure*}[t]
    \centering
    \includegraphics[width=1.0\linewidth]{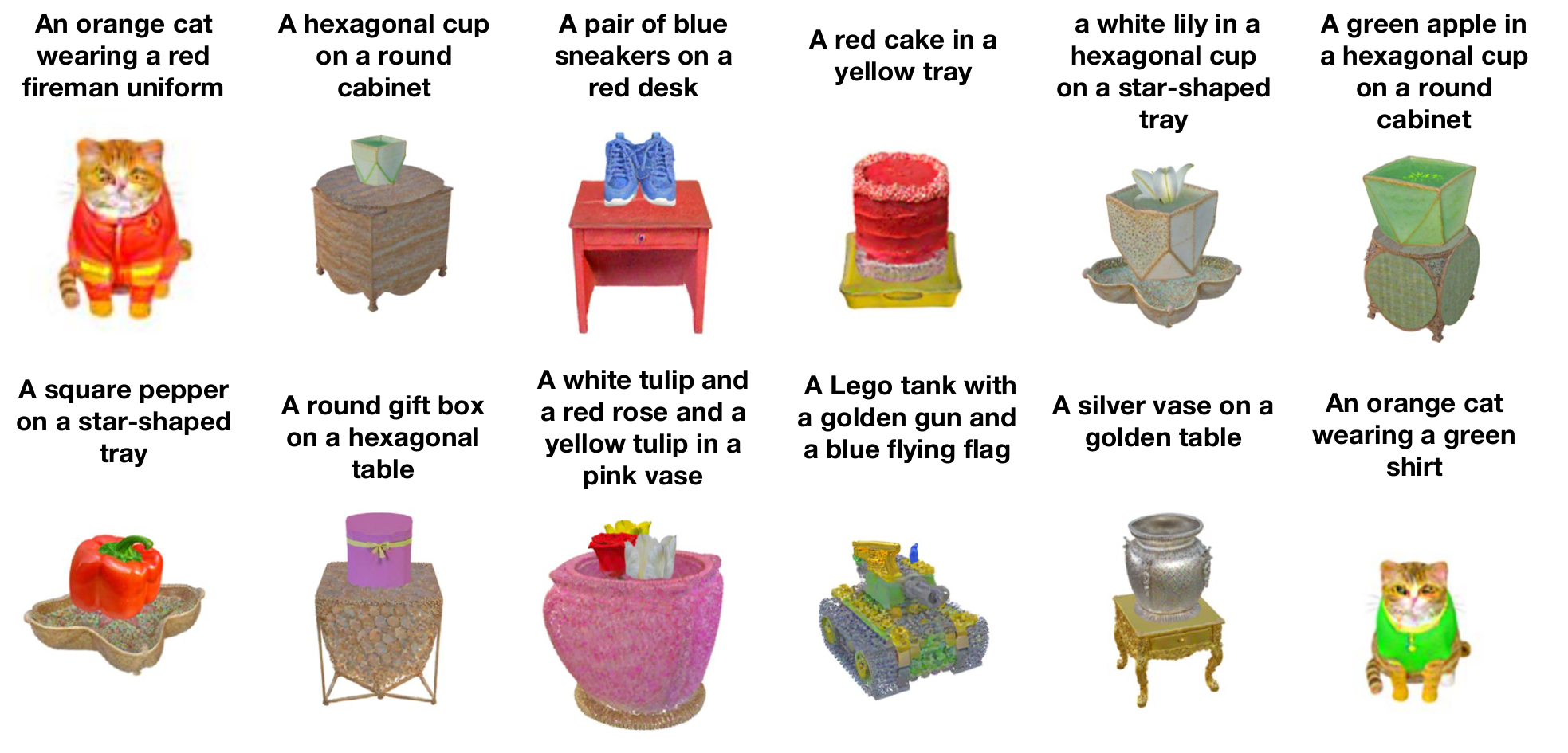}
    \caption{\textbf{Visualization on CSP-100 dataset.} These results are from HCoG based on GALA3D. 
    }
    \label{fig:suppl-exp}
\end{figure*}

\begin{figure*}[t]
    \centering
    \includegraphics[width=1.0\linewidth]{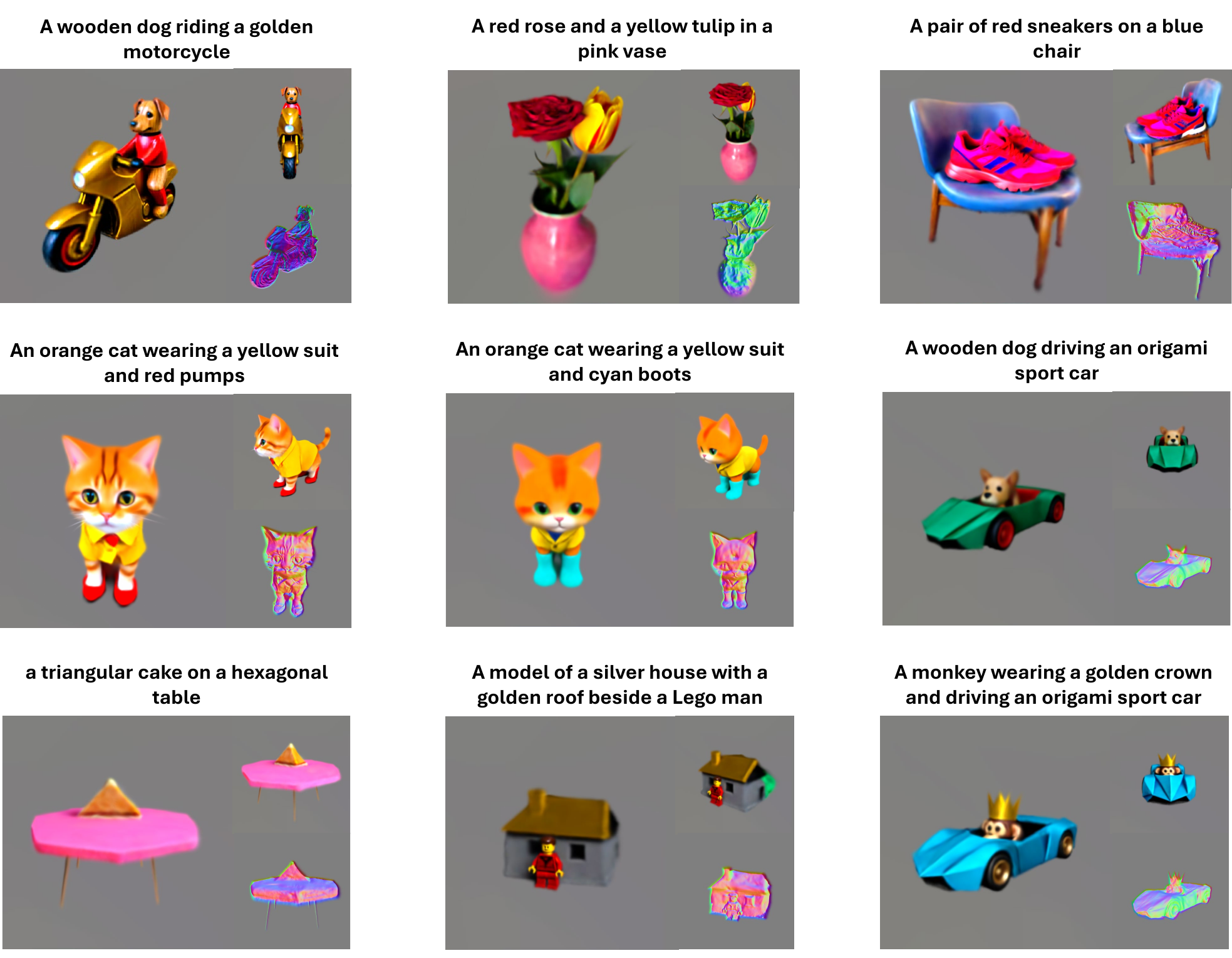}
    \caption{\textbf{Visualization on CSP-100 dataset.} These results are from HCoG based on GaussianDreamer equipped with Stable Diffusion v3.}
    \label{fig:suppl-exp-2}
\end{figure*}

\section{More Experiments}
\label{sec:suppexp}

\subsection{Experiments on CSP-100 Test Dataset}
\label{sec:suppl-csp}


We conduct additional experiments on the CSP-100 dataset introduced by~\cite{Cheng2023Progressive3DPL}. This dataset consists of 100 input texts with high compositional complexity, serving as a challenging benchmark for evaluating the model’s ability to interpret and generate content based on richly attributed textual descriptions. Most of the texts involve multiple entities or instances, each associated with distinct attributes. To assess performance on this dataset, we report both the BLIP-VQA~\cite{Huang2023T2ICompBenchAC} Score and the CLIP Score~\cite{Hessel2021CLIPScoreAR} for quantitative comparison.
Unlike the main text's BLIP-VQA metric of averaging part-wise scores, we adhere to the original metric of multiplying part-wise scores. This adjustment accounts for the fact that CSP-100 input texts are significantly simpler than the cases analyzed in the main paper.

\begin{table}[t]
\centering
\resizebox{1.0\linewidth}{!}{\begin{tabular}{lccc}
\hline
            & Progressive3D~\cite{Cheng2023Progressive3DPL} & HCoG & HCoG+SD3 \\ \hline
BLIP-VQA Score  & 0.474 & 0.518 & \textbf{0.566} \\
CLIP Score        & 29.2 & 29.3 & \textbf{29.7} \\ \hline
\end{tabular}}
\caption{\textbf{Quantitative comparison on CSP-100 dataset.} HCoG denotes our method based on GALA3D and HCOG+SD3 denotes our method based on GaussianDreamer equipped with Stable Diffusion v3.
}
\label{table:suppl-csp}
\end{table}



The CSP-100 dataset consists of four categories of input texts: \textbf{Color}, \textbf{Shape}, \textbf{Material}, and \textbf{Composition}. The first three categories represent individual object attributes, while the \textbf{Composition} category captures multi-object interactions, where each object is associated with distinct attributes. As shown in Tab.~\ref{table:suppl-csp}, we conduct a \textbf{quantitative comparison} between our method and Progressive3D on the CSP-100 benchmark. Furthermore, we assess the performance of our approach across the four attribute categories, reporting both BLIP-VQA Scores and CLIP Scores in Tab.~\ref{tab:suppl-category}. The results demonstrate that our method achieves the highest performance on complex color descriptions and remains highly competitive across the shape, material, and composition categories.

\begin{table}[h]
\centering
\begin{tabular}{|c|c|c|c|}
\hline
  & Category & BLIP-VQA & CLIP Score \\ \hline
\multirow{5}{*}{HCoG} & color & \textbf{0.653} & \textbf{29.7} \\ \cline{2-4}
                             & shape & 0.438 & 28.0 \\ \cline{2-4}
                             & material & 0.562 & \textbf{29.7} \\ \cline{2-4}
                             & composition & 0.488 & 29.4 \\ \cline{2-4}
                             & total & 0.518 & 29.3 \\ \hline
\multirow{5}{*}{HCoG+SD3} & color & \textbf{0.676} & \textbf{30.1} \\ \cline{2-4}
                             & shape & 0.479 & 29.2 \\ \cline{2-4}
                             & material & 0.615 & 29.9 \\ \cline{2-4}
                             & composition & 0.412 & 29.3 \\ \cline{2-4}
                             & total & 0.566 & 29.7 \\ \hline
\end{tabular}
\caption{\textbf{Quantitative results of our method on CSP-100 dataset.} We tested the results of different categories of attributes binding, and our method is best at dealing with \textbf{color} attributes.}
\label{tab:suppl-category}
\end{table}

Additionally, we visualize some results of our method on CSP-100 dataset, which are shown in Fig.~\ref{fig:suppl-exp} and Fig.~\ref{fig:suppl-exp-2}. Fig.~\ref{fig:suppl-exp} shows the results from HCoG based on GALA3D and Fig.~\ref{fig:suppl-exp-2} shows the results from HCoG based on GaussianDreamer equipped with Stable Diffusion v3. 
Our method performs relatively well on CSP-100 dataset in most cases.

\begin{figure*}[h]
    \centering
    \includegraphics[width=0.95\linewidth]{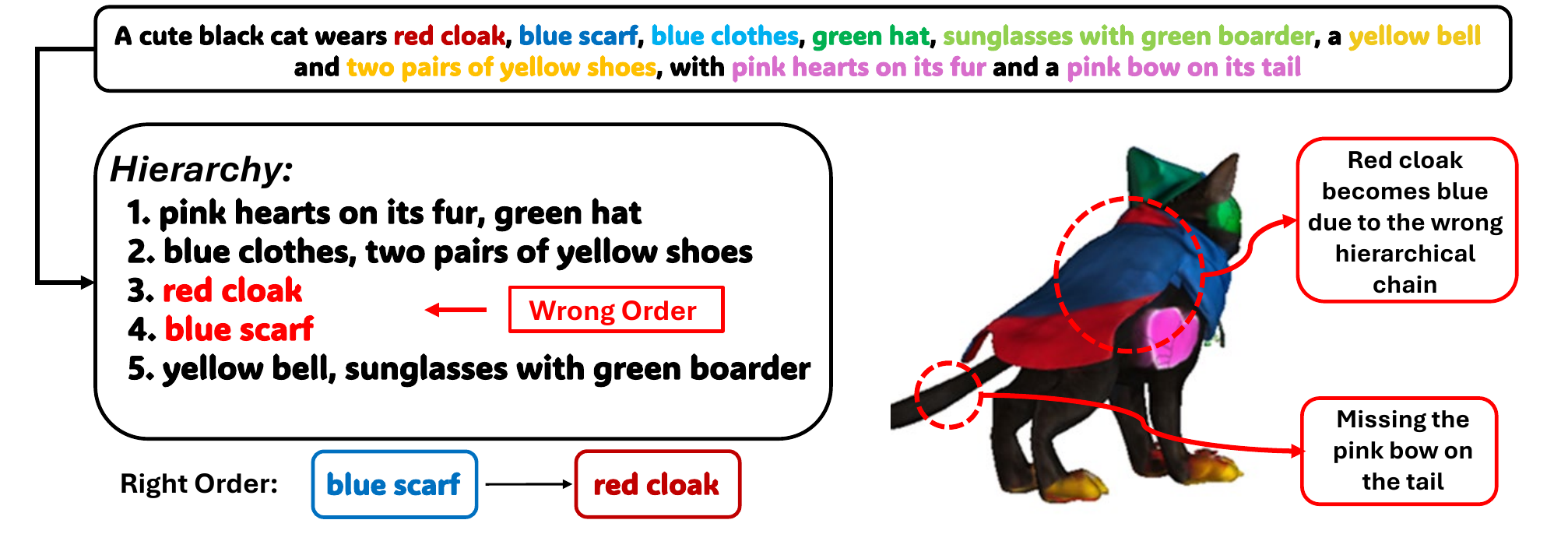}
    \caption{\textbf{Failure case for hierarchy chain and results.} There is a wrong order between red cloak and blue scarf, and the pink bow is missing.}
    \label{fig:re-fail}
\end{figure*}

\subsection{Reliability Analysis of Large Language Model}
\label{sec:analysisofLLM}

Because our method relies on the LLM's ability to generate a
sequential order, we examined its reliability through the following experiment. Specifically, we selected 300 text prompts with varying part counts $n$ (where higher $n$ indicates greater complexity) and deliberately introduced occlusion between parts. We then used the LLM to produce hierarchical chains. Eleven human experts evaluated the accuracy of the extracted parts (\textit{Num Acc}) and the plausibility of the hierarchical chain (i.e., part ordering based on occlusion) (\textit{Chain Acc}). The experts are also guided to create their own hierarchical chains, which were used to compute the number of inversions in the LLM's output (\textit{Inversion}), as shown in Table~\ref{tab:re}.

The LLM achieves consistently high part and chain accuracy ($\geq 95\%$), with few inversions, suggesting its robustness in constructing a proper generation order. However, the LLM may fail when $n$ is large. One such failure is illustrated in Figure~\ref{fig:re-fail}, where an incorrect chain order and missing parts alter the corresponding attributes, leading to misaligned results. The correct generation order should be the scarf first, followed by the cloak. However, the LLM-generated order is reversed, generating the cloak before the scarf. This causes the outer-layer cloak to be influenced by the subsequent scarf generation, resulting in the red cloak being incorrectly rendered as blue. Additionally, the pink bow on the cat's tail is missing. Under similarly complex textual inputs, such failure cases are likely to occur with the LLM.

\begin{table}[h]
    \centering
    \renewcommand{\arraystretch}{0.85} 
    \setlength{\tabcolsep}{8pt} 
    \resizebox{0.99\linewidth}{!}{
    \begin{tabular}{lccc} 
        \toprule
        \textbf{} & \textbf{$1 \leq n \leq 5$} & \textbf{$6 \leq n \leq 10$} & \textbf{$n \geq 11$} \\
        \midrule
        \textbf{Num Acc $\uparrow$}  & \textbf{1.00} & 0.99 & 0.98 \\
        \textbf{Chain Acc $\uparrow$}  & \textbf{0.99} &  0.97 & 0.95 \\
        \textbf{Inversion $\downarrow$}  & \textbf{0.95} &  2.00 & 3.65 \\
        \bottomrule
    \end{tabular}}
    \caption{\textbf{LLM ability for the hierarchy of parts and chain orders.} In most cases, LLM can achieve satisfactory performance.}
    \label{tab:re}
\end{table}

\subsection{Long-CLIP}
\label{sec:supply-longclip}

\begin{figure}[ht]
    \centering
    \includegraphics[width=1.0\linewidth]{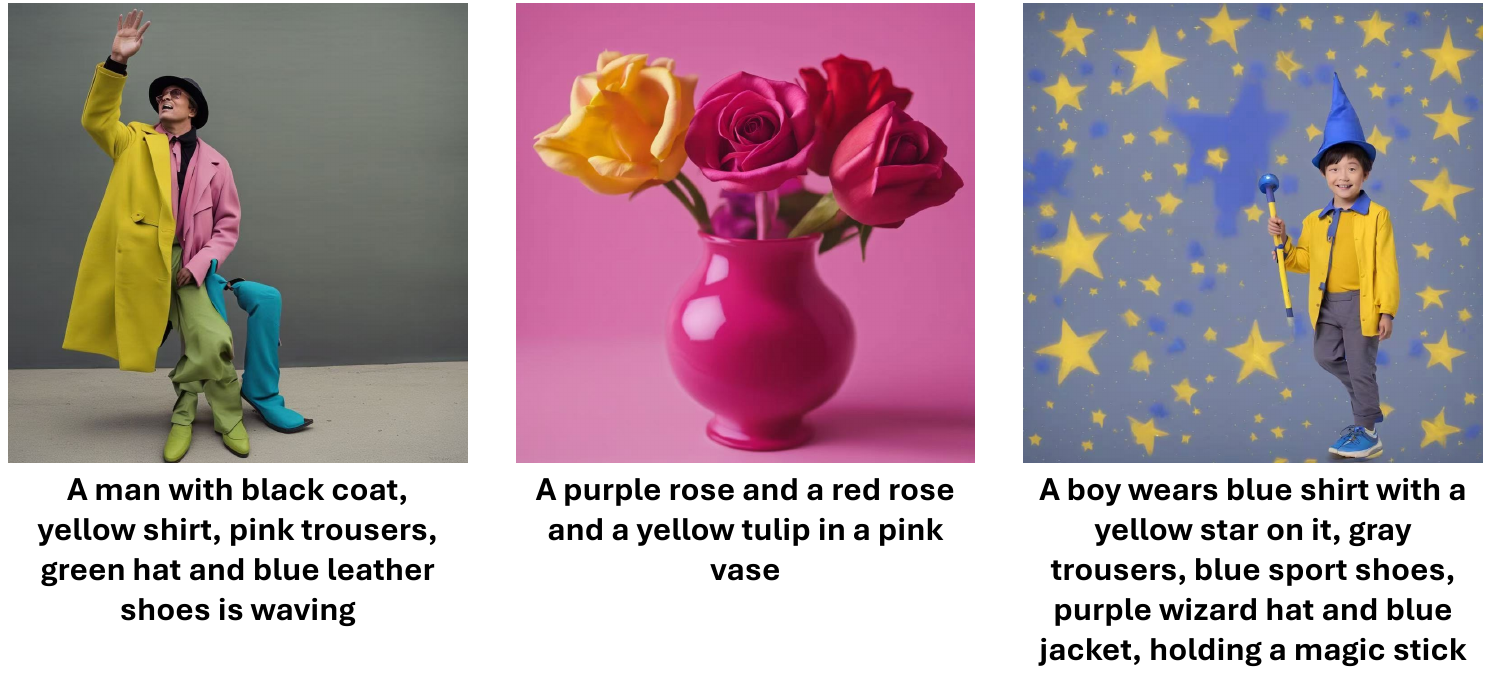}
    \caption{\textbf{The results of diffusion model with Long-CLIP in terms of complex input text.} Diffusion model with Long-CLIP is still unable to solve the problem of complex attributes binding.}
    \label{fig:suppl-longclip}
\end{figure}

We testify other approaches to handle the challenge of long and complex input text, one potential solution is to employ a CLIP variant capable of handling such inputs, such as Long-CLIP~\cite{Zhang2024LongCLIPUT}. We therefore utilize Long-CLIP~\cite{Zhang2024LongCLIPUT} to generate images from long and complex textual descriptions, as illustrated in Fig.~\ref{fig:suppl-longclip}. However, experimental results show that even with Long-CLIP, handling complex attribute-rich text remains challenging in 2D image generation. Since 3D generation builds upon 2D representations, directly applying Long-CLIP to 3D generation tasks yields limited improvement.





\end{document}